\documentclass{article} 
\usepackage[dvipsnames]{xcolor}
\usepackage{sigbovik2019_conference,times}
\usepackage[hidelinks]{hyperref}
\usepackage{url}
\usepackage{soul}
\usepackage{dirtytalk}
\usepackage{epigraph}
\usepackage{subfig}
\usepackage{pifont}
\usepackage{amssymb}

\usepackage{listings}
\usepackage{amsmath}
\usepackage{graphicx}
 \newcommand{\methodName}{SOAR}
 \newcommand{\sota}{SotA}

\usepackage{pgfplots}
\pgfplotsset{width=15cm, height=7cm,compat=1.8}

\definecolor{codegreen}{rgb}{0,0.6,0}
\definecolor{codegray}{rgb}{0.5,0.5,0.5}
\definecolor{codepurple}{rgb}{0.58,0,0.82}
\definecolor{backcolour}{rgb}{0.95,0.95,0.92}

\lstdefinestyle{pystyle}{
    backgroundcolor=\color{backcolour},   
    commentstyle=\color{codegray},
    keywordstyle=\color{codegreen},
    numberstyle=\tiny\color{codegray},
    stringstyle=\color{codepurple},
    basicstyle=\ttfamily\footnotesize,
    breakatwhitespace=false,         
    breaklines=true,                 
    captionpos=b,                    
    keepspaces=true,                 
    numbers=left,                    
    numbersep=5pt,                  
    showspaces=false,                
    showstringspaces=false,
    showtabs=false,                  
    tabsize=2
}

\title{State-of-the-Art Reviewing: a radical\\proposal to improve scientific publication}

\author{Samuel Albanie, Jaime Thewmore, Robert McCraith, Joao F. Henriques \\
SOAR Laboratory,\\
Shelfanger, UK\\
}

\iclrfinalcopy 

\begin{document}
\maketitle


\begin{abstract}
Peer review forms the backbone of modern scientific manuscript evaluation.
But after two hundred and eighty-nine years of egalitarian service to the scientific community, does this protocol remain fit for purpose in 2020? In this work, we answer this question in the negative (strong reject, high confidence) and propose instead \textit{State-Of-the-Art Review} (\methodName), a neoteric reviewing pipeline that serves as a \say{plug-and-play} replacement for peer review.  At the heart of our approach is an interpretation of the review process as a multi-objective, massively distributed and extremely-high-latency optimisation, which we scalarise and solve efficiently for PAC and CMT-optimal solutions.

We make the following contributions: (1) We propose a highly scalable, fully automatic methodology for review, drawing inspiration from best-practices from premier computer vision and machine learning conferences; (2) We explore several instantiations of our approach and demonstrate that \methodName{} can be used to both review prints and pre-review pre-prints; 
(3) We wander listlessly in vain search of catharsis from our latest rounds of savage CVPR rejections\footnote{W.A/W.A/B $\rightarrow$ Reject. A single heavily caffeinated tear, glistening in the flickering light of a faulty office desk lamp, rolls down a weary cheek and falls onto the page. The footnote is smudged. The author soldiers on.}. 

\end{abstract}

\epigraph{If a decision tree in a forest makes marginal improvements, and no one is around to publish it, is it really ``state-of-the-art''?}{George Berkeley,\\\emph{A Treatise Concerning the Principles of Human Knowledge (1710)}}

\section{Introduction} \label{sec:intro}

The process of \textit{peer review}---in which a scientific work is subjected to the scrutiny of experts in the relevant field---has long been lauded an effective mechanism for quality control.  Surgically inserted into the medical field by the cutting-edge work of \citep{adab-al-tabib}, it ensured that treatment plans prescribed by a physician were open to criticism by their peers. Upon discovery of a lengthy medical bill and a dawning realization that theriac was not the \say{wonder drug} they had been promised, unhappy patients could use these \say{peer reviews} as evidence in the ensuing friendly legal proceedings.

Despite this auspicious start, it took many years for the peer review protocol to achieve the popular form that would be recognised by the layperson on the Cowley Road omnibus today.  Credit for this transformation may be at least partially attributed to the Royal Society of Edinburgh who were among the first to realise the benefits of out-sourcing complex quality assessments to unpaid experts~\citep{spier2002history}.

Effacing the names of these heroic contributors, in a process euphemistically called \emph{anonymous review}, was a natural progression.
Attempts to go further and have the reviewers retroactively pay for the privilege of reading a now-copyrighted manuscript (at the discounted price of £50)
 somehow did not catch on, despite the publishers' best intentions.
Peer review (not to be confused with the French tradition of Pierre review, or indeed the spectacle of a pier revue) has since gone from strength-to-strength, and is now the primary quality filtration system for works of merit in both the scientific and TikTok communities.

\begin{figure}
    \centering
    \includegraphics[width=\textwidth]{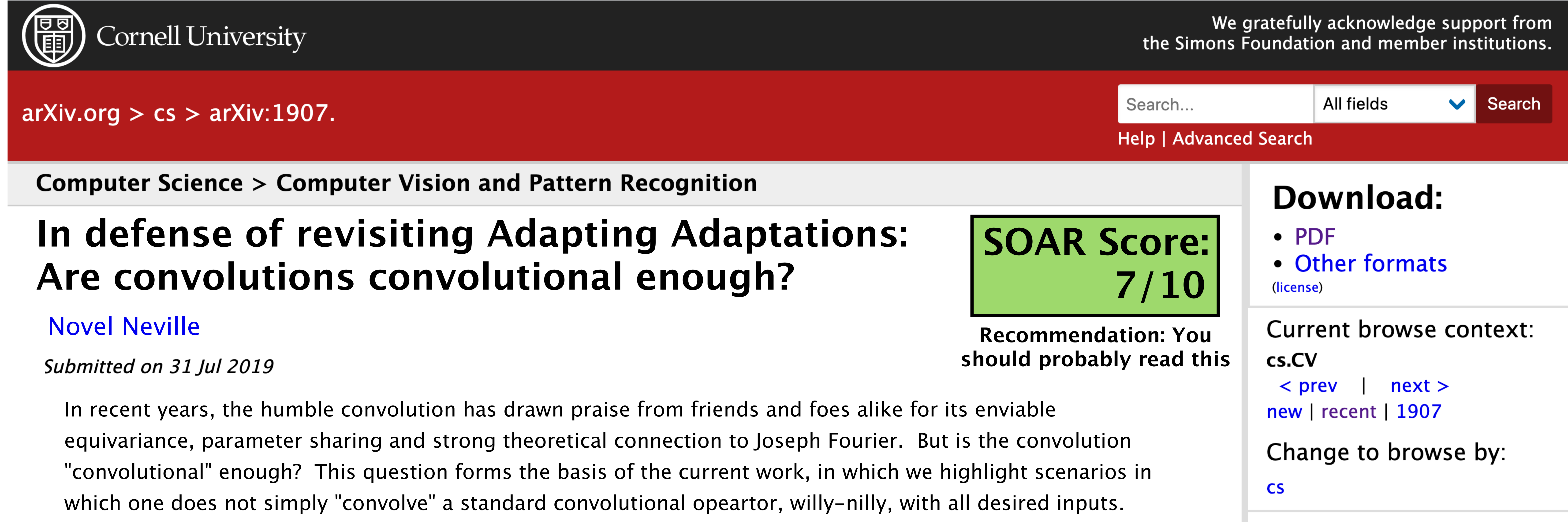}
    \caption{\textbf{Proposed arXiv-integration}: The arXiv server is an invaluable resource that has played a critical role in the dissemination of scientific knowledge.  Nevertheless, a key shortcoming of the current implementation is that it is \textit{unopinionated}, and offers little guidance in whether to invest time in reading each article.  The \methodName{} plugin takes a different approach: summarising the scientific value of the work as an easily digestible score (out of ten) and offering a direct read/don't read recommendation, saving the reader valuable time.
    Future iterations will focus on removing the next bottleneck, the time-consuming ``reading'' stage.}
    \label{fig:arxiv-plugin}
\end{figure}

Still, something is rotten in the state of reviewing. To determine what exactly is causing the smell, our starting point in this work is a critical review of peer review.  We begin by highlighting three key shortcomings of the existing system.

\textbf{Ability to Scale.} As anyone who has prepared for a tech internship interview knows, scale is important.  And so are corner cases.  And so is good communication. But the greatest of these is scale.  To avoid carelessly ruling out future careers at Google, we therefore demonstrate an appreciation of the critical importance of this phenomenon. Indeed, it is here that we must mount our first attack on peer review: its inconvenient linear scaling.  To concretise the implications of this runtime complexity, consider the nation of the United Kingdom which is approximately in Europe.   There are, at the time of writing, 814 hereditary peers in the UK who are born directly into reviewership. Of these, 31 are dukes (7 of which are royal dukes), 34 are marquesses, 193 are earls, 112 are viscounts, and 444 are barons.  

\textbf{Speed.} The mean age of the House of Lords was 70 in 2017.  With a lack of young whippersnappers amidst their ranks, how can we expect these venerable statesmen and stateswomen to do the all-nighters required to review ten conference papers when they are only reminded of the deadline with two days notice because of a bug in their self-implemented calendar app?  One solution is to ensure that they take care when sanitising date/time inputs across time-zones. But even a reliable calendar implementation offers limited defence against a surprise 47 page appendix of freshly minted mathematical notation. The proof of why this is problematic is left as an exercise for the reader.

\textbf{Consistency.} The grand 2014 NeurIPS review experiment~\citep{lawrence2015} provides some insight into the consistency of the peer review process.  When a paper was assigned to two independent review committees, about 57\% of the papers accepted by the first committee were rejected by the second one and vice versa~\citep{price2014}.  While these numbers provide a great deal of hope for anyone submitting rushed work to future editions of the conference, it is perhaps nevertheless worth observing that it brings some downsides.  For one thing, it places great emphasis on the role of registering at the right time to get a lucky paper ID. This, in turn, leads to a great deal of effort on the part of the researcher, who must then determine whether a given ID (for example 5738\footnote{Thankfully, numerology is on hand to supply an answer. \say{5738: You are a step away from the brink that separates big money from lawlessness. Take care, because by taking this step, you will forever cut off your ways to retreat. Unless it is too late.} \citep{5738}}) is indeed, a lucky number, or whether they are best served by re-registering.
A similar phenomenon is observed in large-scale deep learning experiments, which generally consist of evaluating several random initialisations, a job that is made harder by confounders such as hyper-parameters or architectural choices.

By examining the points above, we derive the key following principle for review process design.  \textit{Human involvement---particularly that of elderly British hereditary peers---should be minimised in the modern scientific review process}. In this work, we focus on a particular instantiation of this principle, State-Of-the-Art Reviewing (\methodName), and its mechanisms for addressing these limitations. 

The remainder of the work is structured as follows. In Sec.~\ref{sec:related}, we review related work; in Sec.~\ref{sec:method}, we describe \methodName{}, our bullet-proof idea for automatic reviewing; in Sec.~\ref{sec:implemetation} we develop a practical implementation of the \methodName{} framework, suitable for popular consumption.  Finally, in Sec.~\ref{sec:conclusion}, we conclude with our findings and dreams for swift community adoption.

\begin{figure}
    \centering
    \includegraphics[height=5cm]{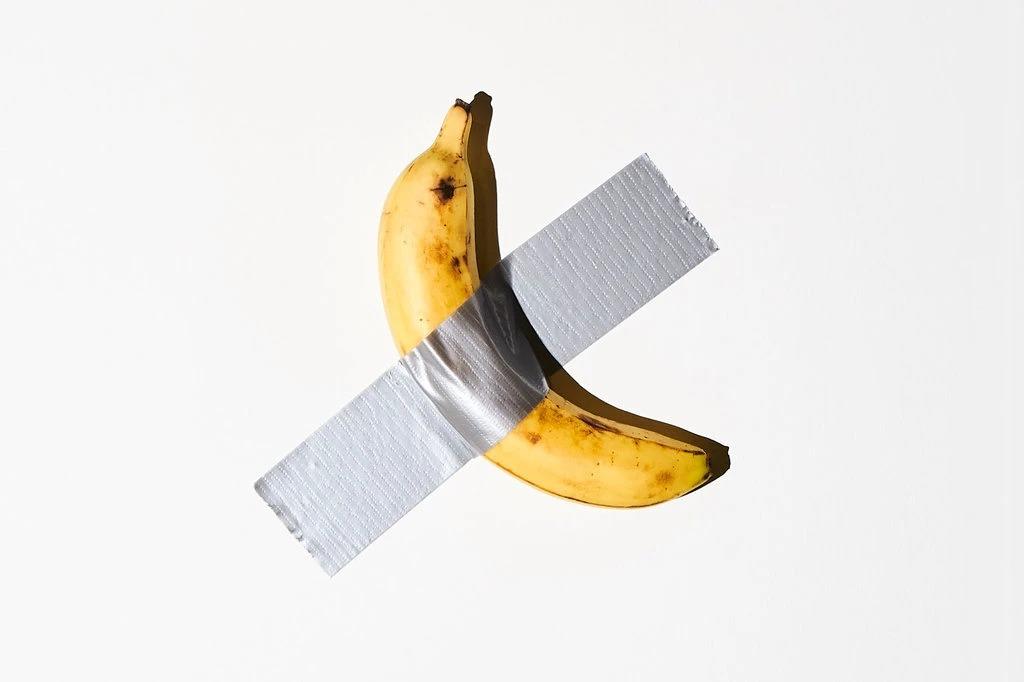}
    \includegraphics[height=5cm]{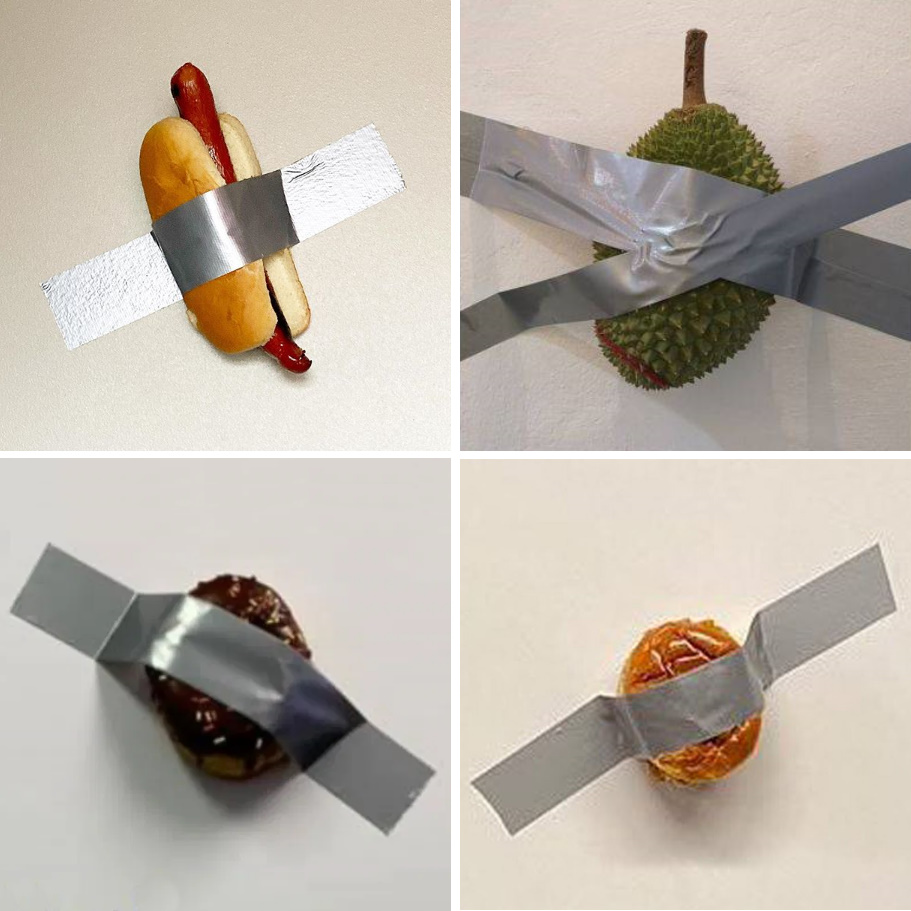}
    \caption{\textbf{(Left)} The state-of-the-art according to \cite{banana},
    \textbf{(Right)} Some marginal improvements by various authors, with questionable added artistic and nutritional value (as measured in calories and milligrams of potassium).\protect\footnotemark}
    \label{fig:banana}
\end{figure}

\footnotetext{Photo credits: (left): \cite{nyt:banana} (top-centre): \cite{hotdog}, (top-right): \cite{durian}, (bottom-center): \cite{donut}, (bottom-right): \cite{PopEyes}}

\section{Related Work \label{sec:related}}

\subsection{Interest in the state-of-the-art} 

Since the discovery of art (Blombos Cave Engravings, ca. 70000 BC) there has been a rising interest in this form of expression, and consequently, the state thereof. From the Medici family of Florence to theatre buff King James I, much effort has been dedicated to patronage of the arts, and much prestige associated with acquiring the latest advances. Pope Julius II was keen to raise the papal state of the art to new heights, namely the ceiling, enlisting the help of renaissance main man Michelangelo. The score of Sistine remains competitive in chapel-based benchmarks, and Michelangelo became a testudine martial artist (with the help of his three equally-talented brothers)~\citep{ninja-turtles}.

From early on, the importance of adding depth was appreciated (Studies on perspective, Brunelleschi, 1415), which continues to this day~\citep{he2016deep}.  Recently, the critically acclaimed work of~\cite{crowley2014state} illustrated how the state-of-the-art can be used to assess the state of art, calling into question the relevance of both hyphens and definite articles in modern computer vision research.  Of least relevance to our work, Fig.~\ref{fig:banana} depicts state-of-the-art developments in the art world. 

\subsection{Literature Review}
\paragraph{The Grapes of Wrath.} In this classic portrayal of the American Dust Bowl, Steinbeck captures the extremes of human despair and oppression against a backdrop of rural American life in all its grittiness. A masterpiece. \ding{72}\ding{72}\ding{72}\ding{72}\ding{72}

\paragraph{Flyer for (redacted) startup, left on a table at NeurIPS 2019 next to a bowl of tortillas.} Hastily put together in PowerPoint and printed in draft-mode amid the death throes of an ageing HP printer, this call for ``dedicated hackers with an appetite for Moonshots, ramen noodles and the promise of stock options'' comes across slightly desperate. \ding{72}\ding{72}

\section{Method} \label{sec:method}


\epigraph{Science is often distinguished from other domains of human culture by its progressive nature: in contrast to art, religion, philosophy, morality, and politics, there exist clear standards or normative criteria for identifying improvements and advances in science.}{\emph{Stanford Encyclopedia of Philosophy}}

In Sec.~\ref{sec:intro}, we identified three key weaknesses in the peer review process: (1) inability to scale; (2) slow runtime and (3) inconsistent results. In the following, we describe the \methodName{} review scheme which seeks to resolve each of these shortcomings, and does so at minimal cost to the taxpayer or ad-funded research lab, enabling the purchase of more GPUs, nap-pods and airpods.

\begin{figure}
    \centering
    \includegraphics[width=\textwidth]{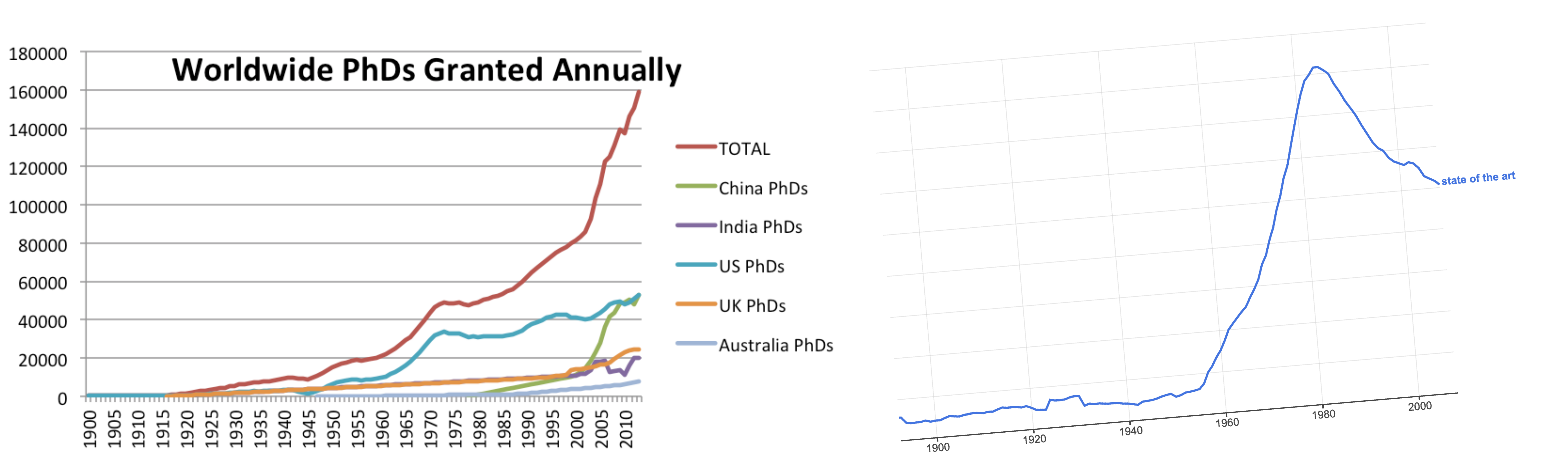}
    \caption{(\textbf{Left}) The number of PhDs granted annually exhibits exponential growth (figure reproduced from ~\cite{gastfriend}), (\textbf{Right}) Google retrieved ngram counts of ``State of the Art'' over the past 200 years of literature. Note that even when the axes are rotated slightly, it remains difficult to preserve an upwards trend. This evidence suggests that either PhDs are becoming exponentially less productive than their predecessors or that the existing reviewing system does not provide sufficient incentivise to use the term ``state-of-the-art'' in modern manuscripts. Our proposal directly addresses the latter.}
    \label{fig:sota-ngram}
\end{figure}

\subsection{State-of-the-art Reviewing (SOAR)}

It is well known is that the quality of a scientific work can be judged along three axes: \textit{efficacy}, \textit{significance} and \textit{novelty}.  Our key insight is that each of these factors can be measured automatically.

\textbf{Assessing efficacy.}  Efficacy is best assessed by determining if the proposed method achieves a new \sota{} (State-of-the-Art). Thankfully, from an implementation perspective, the authors can be relied upon to state this repeatedly in the text.  Thus, rather than parsing results table formats (an error-prone process involving bold fonts and asterisks), we simply word count the occurrences of \say{state-of-the-art} (case insensitive) in the text.  It stands to reason that a higher \sota{} count is preferable.  Moreover, such an approach avoids the embarrassment of realising that one cannot remember what kind of statistical significance test should be applied. 

\textbf{Assessing significance.}  Significance is measured by efficacy. Thus, the efficacy term is weighted twice in the formula.

\textbf{Assessing novelty.}  The assessment of novelty requires close familiarity with prior art and an appreciation for the relative significance of ideas.  We make the key observation that the individuals best placed to make this judgement are the author themselves since they have likely read at least one of the works cited in the bibliography.  We further assume that they will convey this judgement by using the word \say{novel} throughout the document in direct proportion to the perceived novelty of the work. 

With the strategies defined above, we are now in a position to define the SOAR score as follows.

\begin{align}
    \text{SOAR Score} \triangleq \sqrt[3]{S_{\text{\sota}} \cdot S_{\text{\sota}} \cdot S_{\text{novelty}}} \mbox{\text{   /10}}. \label{eqn:soar}
\end{align}

Here $S_{\text{\sota}}$ and $S_{\text{novelty}}$ represent the total occurrences in the manuscript of the terms \say{state-of-the-art} and \say{novel}, respectively.  In both cases, we exclude the related work section (it is important to avoid assigning \sota/novelty credit to the paper under review simply because they cite \sota/novel work). A geometric mean is used to trade-off each factor, but note that a paper must be both \sota{} and novel to achieve a positive SOAR score.  Lastly, we attach a suffix string \say{/10} to every SOAR score for additional credibility.

Note that several factors are \textit{not} assessed: vague concepts like \say{mathematical proofs} and \say{insights} should be used sparingly in the manuscript and are assigned no weight in the review process. If the proof or insight was useful, the authors should use it to improve their numbers.  \sota{} or it didn't happen.

A key advantage of the \methodName{} formula is that it renders explicit the relationship between the key scientific objective (namely, more State-of-the-Art results) and the score.  This lies in stark contrast to peer review, which leaves the author unsure what to optimise.  Consider the findings of Fig.~\ref{fig:sota-ngram}: we observe that although the number of PhDs granted worldwide continues to grow steadily, usage of the term \say{State-of-the-Art} peaked in the mid 1980's.  Thus, under peer review, many PhD research hours are invested every year performing work that is simply not on the cutting edge of science. This issue is directly addressed by measuring the worthiness of papers by their state-of-the-artness rather than the prettiness of figures, affiliation of authors or explanation of methods.

With an appropriately increased focus on SotA we can also apply a filter to conference submissions to immediately reduce the number of papers to be accepted. 
With top conferences taking tens of thousands of submissions each typically requiring three or more reviewers to dedicate considerable time to perform each review, the time savings over an academic career could be readily combined to a long sabbatical, a holiday to sunny Crete, or an extra paper submission every couple of weeks. \\

\section{Implementation} \label{sec:implemetation}

In this section, we outline several implementations of \methodName{} and showcase a use case.

\subsection{Software Implementation and Complexity Analysis}

We implement the \methodName{} algorithm by breaking the submission into word tokens and passing them through a Python 3.7.2 \texttt{collections.Counter} object. We then need a handful of floating-point operations to produce the scalar component of Eqn.~\ref{eqn:soar}, together with a string formatting call and a concatenation with the \say{$/10$}.  The complexity of the overall algorithm is judged reasonable. 

\subsection{Wetware Implementation and Complexity Analysis}

In the absence of available silicon, \methodName{} scoring can also be performed by hand by an attentive graduate student (GS) with a pencil and a strong tolerance to boredom.  Much of the complexity here lies in convincing the GS that it's a good use of time. Initial trials have not proved promising.

\subsection{arXiv integration}
We apply the \methodName{} scoring software implementation to the content of arXiv papers as a convenient Opera browser plugin.  The effect of the plugin can be seen in Fig.~\ref{fig:arxiv-plugin}: it provides a high-quality review of the work in question.  Beyond the benefits of scalability, speed and consistency, this tool offers a direct \say{read/don't read} recommendation, thereby saving the reader valuable time which can otherwise be re-invested into rejecting reviewer invitations emails to compound its savings effect.  We hope that this \textit{pre-review for pre-prints} model will be of great utility to the research community.

\section{Conclusion} \label{sec:conclusion}

In this work, we found that human-based peer review is a succulent and low-hanging fruit, ripe for job automation. With \methodName{}, we have plucked it from the bush and made scalable breakfast of it.  
In future work, we intend to further optimise our implementation of \methodName{} (from 2 LoC to potentially 1 or 0 LoC, in a ludic exercise of code golf). 
With a clear and interpretable reviewing metric, we hope that this work serves to usher in a golden era of scientific research as authors seek to produce peereto-optimal work that is both very novel and very State-of-the-Art.


\bibliographystyle{iclr2019_conference}
\bibliography{iclr2016_conference}

\begin{thebibliography}{16}
\providecommand{\natexlab}[1]{#1}
\providecommand{\url}[1]{\texttt{#1}}
\expandafter\ifx\csname urlstyle\endcsname\relax
  \providecommand{\doi}[1]{doi: #1}\else
  \providecommand{\doi}{doi: \begingroup \urlstyle{rm}\Url}\fi

\bibitem[Ali~al Rohawi(CE 854-–931)]{adab-al-tabib}
Ish\={a}q~bin Ali~al Rohawi.
\newblock Adab al--tabib (practical ethics of the physician).
\newblock CE 854-–931.

\bibitem[Cattelan et~al.(2020)Cattelan, a~tar-covered seagull, and a~very
  strange trip to the~local 7-Eleven]{banana}
Maurizio Cattelan, a~tar-covered seagull, and a~very strange trip to the~local
  7-Eleven.
\newblock Comedian.
\newblock Art Basel Miami Beach, 2020.
\newblock (presumably also visible while not under the influence of
  psychotropic substances).

\bibitem[Crowley \& Zisserman(2014)Crowley and Zisserman]{crowley2014state}
Elliot~J Crowley and Andrew Zisserman.
\newblock The state of the art: Object retrieval in paintings using
  discriminative regions.
\newblock 2014.

\bibitem[Durian(2019)]{durian}
99~Old~Trees Durian.
\newblock Durian tape to white wall, 2019.
\newblock URL
  \url{https://www.facebook.com/99oldtrees/posts/2575690792538528:0}.
\newblock [Online; accessed 27-March-2020].

\bibitem[Eastman \& Laird(1984)Eastman and Laird]{ninja-turtles}
Kevin Eastman and Peter Laird.
\newblock \emph{Teenage Mutant Ninja Turtles}.
\newblock Mirage Studios, 1984.

\bibitem[Gastfriend(2015)]{gastfriend}
Eric Gastfriend.
\newblock 90\% of all the scientists that ever lived are alive today, 2015.
\newblock URL
  \url{https://futureoflife.org/2015/11/05/90-of-all-the-scientists-that-ever-lived-are-alive-today/?cn-reloaded=1}.
\newblock [Online; accessed 27-March-2020].

\bibitem[He et~al.(2016)He, Zhang, Ren, and Sun]{he2016deep}
Kaiming He, Xiangyu Zhang, Shaoqing Ren, and Jian Sun.
\newblock Deep residual learning for image recognition.
\newblock In \emph{Proceedings of the IEEE conference on computer vision and
  pattern recognition}, pp.\  770--778, 2016.

\bibitem[Lawrence \& Cortes(2015)Lawrence and Cortes]{lawrence2015}
Neil Lawrence and Corinna Cortes.
\newblock Examining the repeatability of peer review, 2015.
\newblock URL
  \url{http://inverseprobability.com/talks/slides/nips_radiant15.slides.html#/}.
\newblock [Online; accessed 27-March-2020].

\bibitem[Noennig(2019)]{hotdog}
Jordyn Noennig.
\newblock Banana duct-taped to the wall is art. but how about sausage taped to
  the wall because, wisconsin, 2019.
\newblock URL \url{https://eu.jsonline.com/story/entertainment/2019/12/1
  0/banana-duct-taped-wall-sparks-vanguard-milwaukees-hot-dog-wall/4384303002/}.
\newblock [Online; accessed 27-March-2020].

\bibitem[numeroscop.net(2020)]{5738}
numeroscop.net.
\newblock On 5738, 2020.
\newblock URL
  \url{https://numeroscop.net/numerology_number_meanings/four_digit_numbers/number_5738.html}.
\newblock [Online; accessed 27-March-2020].

\bibitem[NYT-Photography(2019)]{nyt:banana}
NYT-Photography.
\newblock The 120,000 dollar banana wins art basel, 2019.
\newblock URL
  \url{https://www.nytimes.com/2019/12/06/style/art-basel-miami-beach.html}.
\newblock [Online; accessed 27-March-2020].

\bibitem[Popeyes(2019)]{PopEyes}
Popeyes.
\newblock Chicken taped to wall, 2019.
\newblock URL
  \url{https://twitter.com/popeyeschicken/status/1203140095005605888}.
\newblock [Online; accessed 27-March-2020].

\bibitem[Price(2014)]{price2014}
Eric Price.
\newblock The nips experiment, 2014.
\newblock URL \url{http://blog.mrtz.org/2014/12/15/the-nips-experiment.html}.
\newblock [Online; accessed 27-March-2020].

\bibitem[Spier(2002)]{spier2002history}
Ray Spier.
\newblock The history of the peer-review process.
\newblock \emph{TRENDS in Biotechnology}, 20\penalty0 (8):\penalty0 357--358,
  2002.

\bibitem[Tampa-Police-Department(2019)]{donut}
Tampa-Police-Department.
\newblock Sgt. donut, 2019.
\newblock URL \url{https://www.facebook.com/TampaPD/posts/3297203563685153}.
\newblock [Online; accessed 27-March-2020].

\bibitem[{Wikipedia contributors}(2020)]{wiki:peerage}
{Wikipedia contributors}.
\newblock Hereditary peer --- {Wikipedia}{,} the free encyclopedia, 2020.
\newblock URL
  \url{https://en.wikipedia.org/w/index.php?title=Hereditary_peer&oldid=946076588}.
\newblock [Online; accessed 27-March-2020].

\end{thebibliography}

\appendix

\section{Appendix} 

In the sections that follow, we provide additional details that were carefully omitted from the main paper.

\subsection{Title Pronunciation \label{app:pronunciation}}

\begin{figure}[h]
    \centering
    \includegraphics[width=0.6\textwidth,trim={0cm 0cm 0cm 0cm},clip]{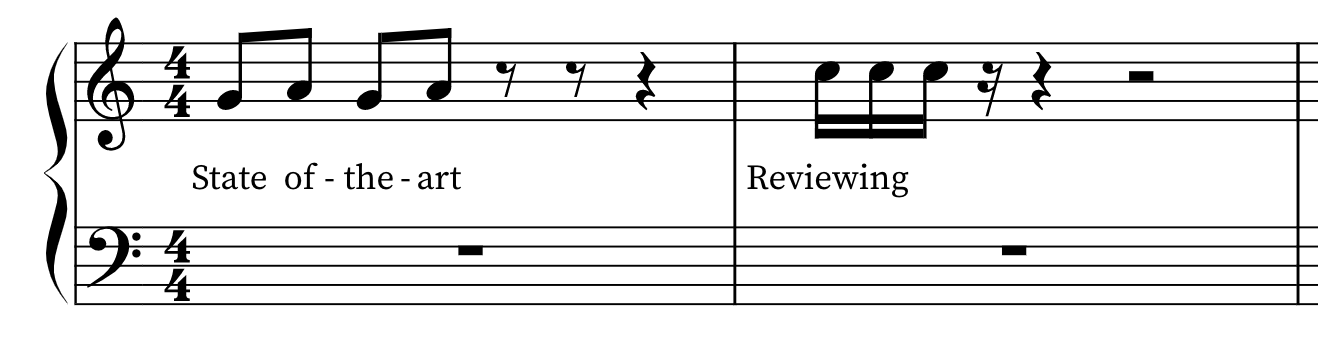}
    \caption{Official Title Pronunciation Guide (gustoso).}
    \label{fig:music-score}
\end{figure}

In common with prior works, we hope that the arguments put forward in this paper will spark useful discussion amongst the community.  Where appropriate, we encourage the reader to use the official title pronunciation guide in Fig.~\ref{fig:music-score}.

\subsection{Inheritance of peerages \label{app:inheritance}}  One historical challenge with the expansion of the UK peerage system has been something of a pre-occupation with preventing the passage of peerages to women.  We note that this has grave implications for the ability to scale peer review.  Consider the progressive rules for inheritance under Henry IV of England~\citep{wiki:peerage}, which were as follows: 

\begin{quote}
\say{If a man held a peerage, his son would succeed to it; if he had no children, his brother would succeed. If he had a single daughter, his son-in-law would inherit the family lands, and usually the same peerage; more complex cases were decided depending on circumstances. Customs changed with time; earldoms were the first to be hereditary, and three different rules can be traced for the case of an Earl who left no sons and several married daughters. In the 13th century, the husband of the eldest daughter inherited the earldom automatically; in the 15th century, the earldom reverted to the Crown, who might re-grant it (often to the eldest son-in-law); in the 17th century, it would not be inherited by anybody unless all but one of the daughters died and left no descendants, in which case the remaining daughter (or her heir) would inherit.}
\end{quote}

Note that by avoiding the necessity of a direct bloodline between peers, \methodName{} neatly sidesteps this scalability concern, further underlining its viability as a practical alternative to traditional peer review.

\subsection{New Insights: A memoryless model for scientific progress}

\begin{figure}[!h]
    \centering
    \includegraphics[width=0.6\textwidth]{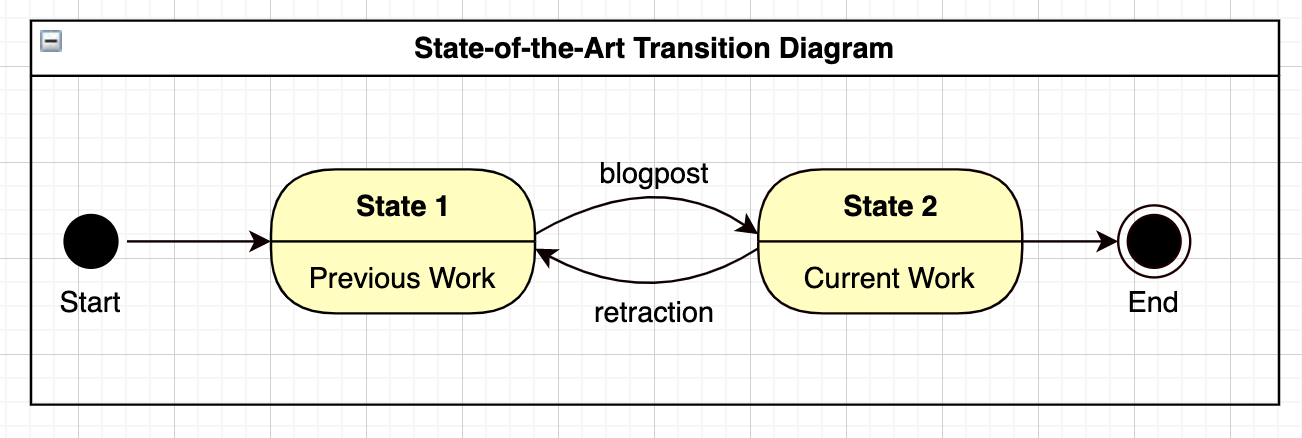}
    \caption{By introducing a State-of-the-Art state transition diagram, we show how the progression of research can be modelled as a memoryless automaton.}
    \label{fig:state-diagram}
\end{figure}

Beyond time savings for reviewers, we note here that the \methodName{} score further provides insights into the scientific method itself, yielding time savings for authors too.  To illustrate this, we provide a state transition diagram in Fig.~\ref{fig:state-diagram} which models the evolution of research progress.  Importantly, this model guarantees a Markov-optimal approach to research: a researcher must only ever read the paper which represents the current State-of-the-Art to make further progress.

\end{document}